\documentclass[11pt,a4paper]{article}
\usepackage[hyperref]{emnlp-ijcnlp-2019}
\usepackage{times}
\usepackage{latexsym}
\usepackage{graphicx}
\usepackage{amsfonts}
\usepackage{multirow}
\usepackage{url}
\usepackage{arydshln}

\aclfinalcopy 

\title{
BioNLP-OST 2019 RDoC Tasks: Multi-grain Neural Relevance Ranking Using Topics and Attention Based Query-Document-Sentence Interactions\\
}

\newcommand*{\affaddr}[1]{#1} 
\newcommand*{\affmark}[1][]{\textsuperscript{#1}}

\renewcommand{\thefootnote}{\fnsymbol{footnote}}
\usepackage{blindtext}
\newcommand\blfootnote[1]{%
\begingroup
\renewcommand\thefootnote{}\footnote{#1}%
\addtocounter{footnote}{-1}%
\endgroup
}

\author{*Yatin Chaudhary\affmark[1,2], 
*Pankaj Gupta
\affmark[1,2], 
Hinrich Sch\"{u}tze\affmark[2]\\
	\affaddr{\affmark[1]Corporate Technology, Machine-Intelligence (MIC-DE), Siemens AG Munich, Germany}\\
	\affaddr{\affmark[2]CIS, University of Munich (LMU) Munich, Germany} \\
	{\tt \{yatin.chaudhary, pankaj.gupta\}@siemens.com}\\
}

\date{}

\begin{document}
\maketitle
\begin{abstract}
  This paper presents our system details and results of participation in the RDoC Tasks of BioNLP-OST 2019.
  Research Domain Criteria (RDoC) construct is a multi-dimensional and broad framework to describe mental health disorders by combining knowledge from \textit{genomics} to \textit{behaviour}.
  Non-availability of RDoC labelled dataset and tedious labelling process hinders the use of RDoC framework to reach its full potential in Biomedical research community and Healthcare industry.
  Therefore, \textit{Task-1} aims at retrieval and ranking of PubMed abstracts relevant to a given RDoC construct and \textit{Task-2} aims at extraction of the most relevant sentence from a given PubMed abstract.
  We investigate (1) attention based supervised neural topic model and SVM for retrieval and ranking of PubMed abstracts and, further utilize BM25 and other relevance measures for re-ranking, (2) supervised and unsupervised sentence ranking models utilizing \textit{multi-view} representations comprising of \textit{query-aware attention-based sentence representation (QAR)}, bag-of-words (BoW) and TF-IDF.
  Our best systems achieved 1st rank and scored 0.86 mAP and 0.58 macro average accuracy in Task-1 and Task-2 respectively.
\end{abstract}

\section{Introduction}
\blfootnote{* : Equal Contribution} 
	The scientific research output of the biomedical community is becoming more sub-domain specialized and increasing at a faster pace. 
	Most of the biomedical domain knowledge is in the form of unstructured text data.
	Natural Language Processing (NLP) techniques such as relation extraction and information retrieval have enabled us to effectively mine relevant information from a large corpus.
	These techniques have significantly reduced the time and effort required for knowledge mining and information extraction from past scientific studies and electronic health reports (EHR).

	Information Retrieval (IR) is the process of retrieving relevant information from an unstructured text corpus, which satisfies a given query/requirement, for example Google search, email search, database search etc.
	This is generally achieved by converting the query and the document collection into an external representation which by preserving the important semantical information can reduce the IR processing time.
	This external representation can be generated using either statistical approach i.e., word counts or distributed semantical approach i.e., word embeddings.
	Therefore, there is a motivation to develop such IR system which can understand the specialized sub-domain language and domain-specific jargon of biomedical domain and assist researchers and medical professionals by effectively and efficiently retrieving most relevant information given a query.

    \begin{table*}[t]
    \begin{center}
			\resizebox{16cm}{!}{%
            \begin{tabular}{c|c|c}
            \textbf{PMID}     & \textbf{RDoC Construct}      & \textbf{PubMed Abstract}  \\
            \hline \hline
            14998902 & Acute\_Threat\_Fear & \begin{tabular}[c]{@{}l@{}}\textit{\underline{Title}}: Mother lowers {\color{blue} glucocorticoid} levels of preweaning rats after {\color{blue} acute threat}.\\ \textit{\underline{Abstract}}: Exposure to a {\color{blue} deadly threat}, an adult male rat, induced the release \\ of {\color{blue} corticosterone} in 14-day-old rat pups. The {\color{blue} endocrine stress response} was \\ decreased when the pups were reunited with their mother immediately after exposure. \\ These findings demonstrate that social variables can reduce the consequences of \\ an {\color{blue} aversive experience}.\end{tabular} \\
            \hline
            21950094 & Sleep\_Wakefulness  & \begin{tabular}[c]{@{}l@{}}\textit{\underline{Title}}: Central mechanisms of {\color{red} sleep-wakefulness cycle}\\ \textit{\underline{Abstract}}: Brief anatomical, {\color{red} physiological and neurochemical} basics of the \\ {\color{red} regulation of wakefulness, slow wave (NREM) sleep} and {\color{red} paradoxical (REM) sleep} \\ are regarded as representing by the end of the first decade of the second millennium.\end{tabular}          
            \end{tabular}}
        \end{center}
        \captionof{table}{RDoC construct - This table shows two PubMed abstracts labelled with two different RDoC construct and PubMed ID (PMID). Highlighted words ({\color{blue} blue} and {\color{red} red}) in each abstract shows content words which together provide the semantic understanding of the corresponding RDoC constructs.}
		\label{table:rdoc_construct_examples}
    \end{table*}

    RDoC Tasks aims at exploring information retrieval (IR) and information extraction (IE) tasks on selected abstracts from PubMed dataset.
    While Task-1 aims to rank abstracts i.e., \textit{coarse granularity}, Task-2 aims to rank sentences i.e., \textit{fine granularity} and hence the term \textit{multi-grain}.
    An RDoC construct combines information from multiple sources like genomics, symptoms, behaviour etc. and therefore, is a much broader way of describing mental health disorders than symptoms based approach.
    Table~\ref{table:rdoc_construct_examples} shows the association between PubMed abstracts and RDoC constructs depending on the semantic knowledge of the highlighted content words.
    Both of these tasks aim in the direction of \textit{ease of accessibility} of PubMed abstracts labelled with diverse RDoC constructs so that this information can reach its full potential and can be of help to biomedical researchers and healthcare professionals.
    
    \section{Task Description and Contributions}
    {\bf RDoc-IR Task-1}: 
    The task aims at retrieving and ranking the PubMed abstracts (within each of the eight clusters) that are relevant for the RDoC construct (i.e, a query) related to the cluster in the abstract appears. The training data consists of abstracts (title + sentences) each annotated with one or more RDoC constructs. 
Test data consists of abstracts without annotation and the goal is to submit a ranked lists of relevant articles for each medical domain RDoC construct. 

{\bf RDoc-IE Task-2}
    The task aims at extracting the most relevant sentence from each PubMed abstract for the corresponding RDoC construct. The input consists of an abstract (title $t$ and sentences $s$) for an RDoC construct $q$. The training data consists of abstracts each annotated with one RDoC construct and the most relevant sentence.
Test data contains abstracts relevant for RDoC constructs and the goal is to submit a list of predicted most relevant sentence for each abstract.

{\bf Our Contributions}: Following are our multi-fold contributions in this paper:

(1) {\bf RDoC-IR Task-1}: We perform document (or abstract) ranking in two steps, first using supervised neural topic model and SVM. Moreover, we have introduced attentions in supervised neural topic model, along with pre-trained word embeddings from several sources.  Then, we re-rank documents using BM25 and similarity scores between query and query-aware attention-based document representation.

Comparing with other participating systems in the shared task, our submission is ranked $1^{st}$ with a mAP score of $0.86$.  

(2) {\bf RDoC-IE Task-2}: We have addressed the sentence ranking task by introducing unsupervised and supervised sentence ranking schemes.  Moreover, we have employed multi-view representations consisting of bag-of-words, TF-IDF and query-aware attention-based sentence representation via enhanced query-sentence interactions.   We have also investigated relevance of title with the sentences and coined ways to incorporate both query-sentence and title-sentence relevance scores in ranking sentences with an abstract.

Comparing with other participating systems in the shared task, our submission is ranked $1^{st}$ with a macro average accuracy of $0.58$. Our code is available at \url{https://github.com/YatinChaudhary/RDoC_Task}.

\begin{figure*}[t]
    	\includegraphics[width=\textwidth]{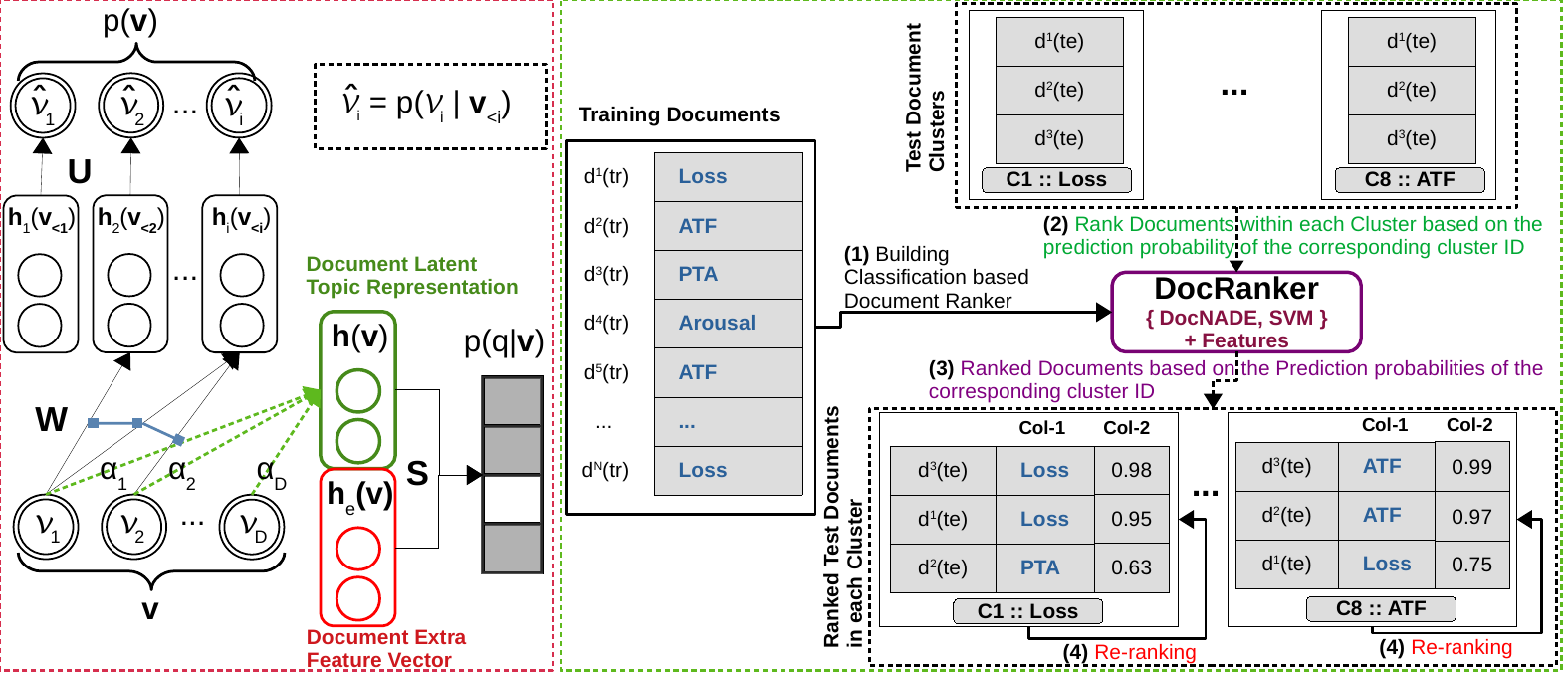}
    	\caption{(Left) DocNADE Topic Model: Blue colored lines signify parameter sharing, $\alpha$ attention weights are used to compute \textit{latent document representation} ${\bf h}({\bf v})$; 
(Right) RDoC Task-1 system architecture, where the numbered arrow (1) denotes the flow. 
 {\bf Col-1} indicates ``predicted label'' by the DocRanker and {\bf Col-2} indicates ``prediction probability'' ($p(q|{\bf v})$). ``Features'' inside DocRanker indicates \textit{FastText} and \textit{word2vec} pretrained embeddings.}
    	\label{fig:task1_pipeline}
    \end{figure*}

\section{Methodology}
	
In this section, we first describe representing a query, sentence and document using local and distributed representation schemes. 
We further describe enhanced query-document (query-title and query-content) and query-sentence interactions to compute query-aware document or sentence representations for Task-1 and Task-2, respectively.   Finally, we discuss the application of supervised neural topic modeling in ranking documents for task 1 and introduce unsupervised and supervised sentence rankers for Task-2.   

     \subsection{Query, Sentence and Document Vectors}\label{sec:textrepresentation}

    In this paper, we deal with texts of different lengths in form of query, sentence and document. 
    In this section, we describe the way we represent the different texts. 

    {\bf Bag-of-words (BoW)} and {\bf Term frequency-inverse document frequency (TF-IDF)}: We use two the local representation schemes: BoW and TF-IDF \cite{DBLP:books/daglib/0021593} to compute sentence/document vectors.  
    
    
  
 {\bf Embedding Sum Representation (ESR)}: Word embeddings \cite{mikolov2013distributed, pennington2014glove} have been successfully used in computing distributed representation of text snippets (short or long).   In ESR scheme, we employ the pre-trained word embeddings from FastText \cite{bojanowski2017enriching} and word2vec \cite{mikolov2013distributed}. To represent a text (query, sentence or document), we compute the sum of (pre-trained) word vectors of each word in the text.  
 E.g.,  ESR for a document $d$ with $D$ words can be computed as: 
      
      $ESR(d) = \widetilde{\bf d} =  \sum_{i=1}^D {\bf e}(d_i)$
where, ${\bf e} \in \mathbb{R}^{E}$ is the pre-trained embedding vector of dimension $E$ for the word $d_i$.  

    

    {\bf Query-aware Attention-based Representation (QAR) for Documents and Sentences}: 
    Unlike ESR, we reward the maximum matches between a query and document by computing density of matches between them, similar to \newcite{mcdonald2018deep}. In doing so, we introduce a weighted sum of word vectors from pre-trained embeddings and therefore, incorporate importance/attention of certain words in document (or sentence) that appear in the query text. 

For an enhanced query-aware attention based document (or sentence) representation, we first compute an  histogram ${\bf a}_{i}(d) \in \mathbb{R}^{D}$ of attention weights for each word $k$ in the document $d$ (or sentence $s$) relative to the $i$th query word $q_i$, using cosine similarity: 
\[
{\bf a}_{i} (d)  =  [a_{i, k}]_{k=1}^{D}  
\mbox{ where, }
a_{i, k} =  \frac{{\bf e}(q_i)^{T} {\bf e}(d_{k})}{||{\bf e}(q_i)||  \ ||{\bf e}(d_{k})||} 
\]
for each $k$th word in the document $d$. Here, ${\bf e}(w)$ refers to an embedding vector of the word $w$.  

We then compute an query-aware attention-based representation $\Phi_i(d)$ of document $d$  from the viewpoint of $i$th query word by summing the word vectors of the document, weighted by their attention scores ${\bf a}_{i} (d)$:
\[
\Phi_i(d) = \sum_{k=1}^{D} a_{i, k} (d) \  {\bf e}(d_k) = {\bf a}_{i} (d) \odot [{\bf e}(d_k)]_{k=1}^{D}
\]
where $\odot$ is an element-wise multiplication operator. 

Next, we compute density of matches between several words in query and the document by summing each of the attention histograms $a_i$ for all the query terms $i$. Therefore, the {\it query-aware document representation} for a document (or sentence) relative to all query words in $q$ is given by:

\begin{equation}\label{eq:QAR}
\mbox{QAR}(d) = \Phi_q (d) =  \sum_{i}^{|q|} \Phi_i(d)
\end{equation}

Similarly, a {\it query-aware sentence representation} $\Phi_q(s)$  and {\it query-aware title representation} $\Phi_q(t)$ can be computed for the sentence $s$ and document title $t$, respectively.  

For query representation, we use ESR scheme as $\widetilde{\bf q} =  \sum_{i=1}^{|q|} {\bf e}(w_i)$. 

\begin{figure*}[t]
    	\includegraphics[width=\textwidth]{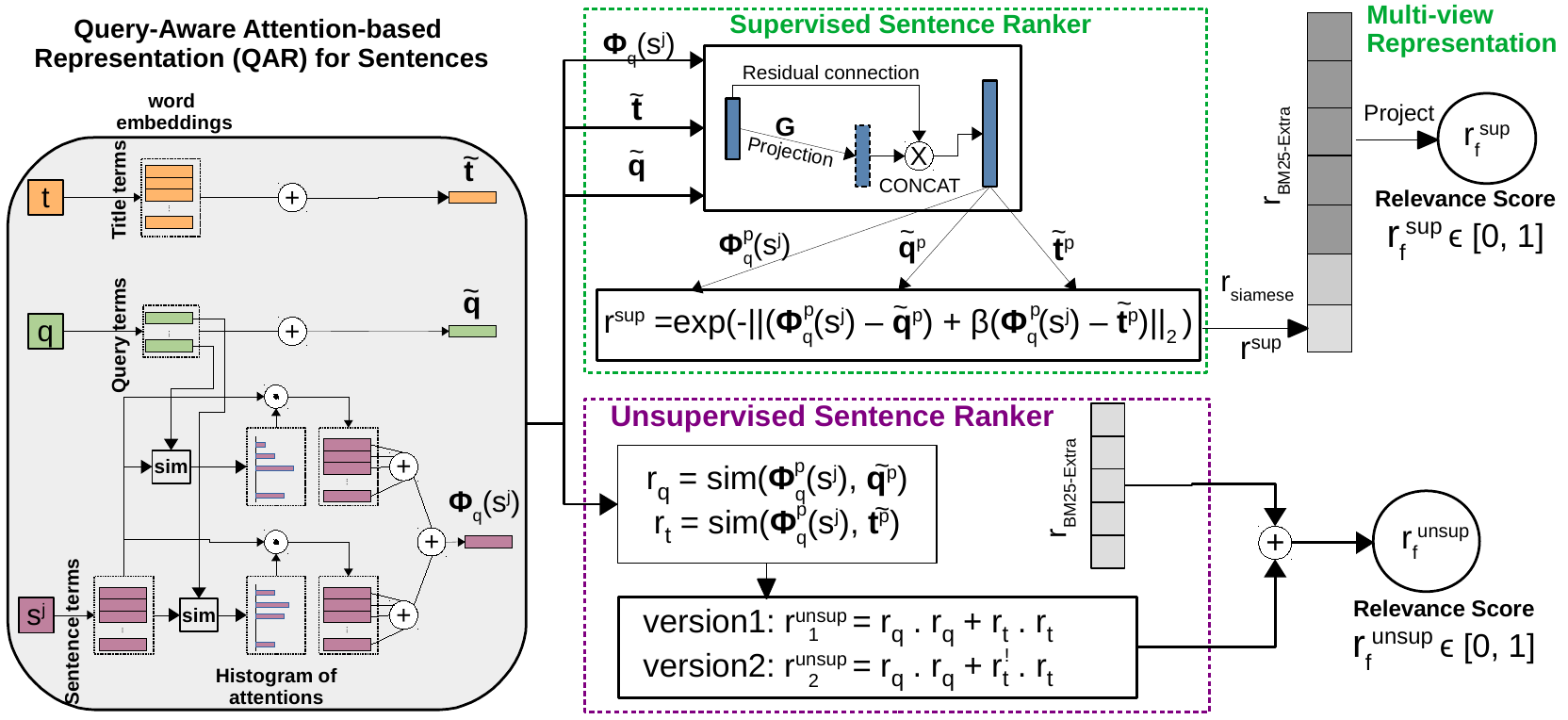}
    	\caption{RDoC Task-2 System Architecture for Supervised and Unsupervised sentence ranking, consisting of: Query-aware Representation, Supervised and Unsupervised sentence rankers for computing the final relevance scores $r_f^{sup}$ and $r_f^{unsup}$, respectively. Here, $sim$ refers to cosine similarity.}
    	\label{fig:task2_pipeline}
    \end{figure*}

Figure~\ref{fig:task2_pipeline} illustrates the computation of {\it query-aware attention-based sentence representation}.

    
    \subsection{Document Neural Topic Models}\label{sec:TM}

 Topic models (TMs)~\cite{blei2003latent} have shown to capture thematic structures, i.e., topics appearing within the document collection. 
Beyond interpretability, topic models can extract latent document representation that is used to perform document retrieval. 
Recently,  \newcite{gupta2019document} and \newcite{gupta2018texttovec} have shown that the neural network-based topic models (NTM) outperform LDA-based topic models \cite{blei2003latent, srivastava2017autoencoding} in terms of generalization, interpretability and document retrieval. 

In order to perform document classification and retrieval, we have employed supervised version of neural topic model with extra features and further introduced word-level attention in a neural topic model, i.e. in DocNADE \cite{larochelle2012neural, gupta2019document}.  

    {\bf Supervised NTM (SupDocNADE)}: 
    Document Neural Autoregressive Distribution Estimator (DocNADE) is a neural network based topic model that works on bag-of-words (BoW) representation to model a document collection in a language modeling fashion. 

    Consider a document $d$, represented as ${\bf v} = [v_1, ..., v_i, ..., v_D]$ of size $D$, where $v_i \in \{1, ..., Z\}$ is the index of $i$th word in the vocabulary and $Z$ is the vocabulary size.
    DocNADE models the joint distribution $p({\bf v})$ of document ${\bf v}$ by decomposing $p({\bf v})$ into autoregressive conditional of each word $v_i$ in the document, i.e., $p({\bf v}) = \sum_{i=1}^D p(v_i|{\bf v}_{<i})$, where ${\bf v}_{<i} \in \{v_1, ..., v_{i-1}\}$.

    As shown in Figure~\ref{fig:task1_pipeline} (left), DocNADE computes each autoregressive conditional $p(v_i|{\bf v}_{<i})$ using a feed forward neural network for $i \in \{1, ..., D\}$ as,
    
\[
p(v_i = w|{\bf v}_{<i}) = \frac{exp(b_w + {\bf U}_{w,:} {\bf h}({\bf v}_{<i}))}{\sum_{w'} exp(b_{w'} + {\bf U}_{w',:} {\bf h}({\bf v}_{<i}))} 
\]
\[
 {\bf h}_i({\bf v}_{<i}) = f({\bf c} + \sum_{j<i} {\bf W}_{:,v_j})
\]
where, $f( \cdot )$ is a non-linear activation function, ${\bf W} \in \mathbb{R}^{H \times Z}$ and ${\bf U} \in \mathbb{R}^{Z \times H}$ are encoding and decoding matrices, ${\bf c} \in \mathbb{R}^{H}$ and ${\bf b} \in \mathbb{R}^{Z}$ are encoding and decoding biases, $H$ is the number of units in latent representation ${\bf h}_i({\bf v}_{<i})$. Here, ${\bf h}_i({\bf v}_{<i})$ contains information of words preceding the word $v_i$. 
    For a document ${\bf v}$, the log-likelihood ${\mathcal L}({\bf v})$ and latent representation ${\bf h}({\bf v})$ are given as,    
\begin{equation}
{\mathcal L}^{unsup}({\bf v}) = \sum_{i=1}^D \log p(v_i|{\bf v}_{<i})
\end{equation}
\begin{equation}\label{eq:embeddingagrregation}
{\bf h}({\bf v}) = f({\bf c} + \sum_{i=1}^D {\bf W}_{:,v_i})
\end{equation}

Here, ${\mathcal L}({\bf v})$ is used to optimize the topic model in unsupervised fashion and ${\bf h}({\bf v})$ encodes the topic proportion. 
See \newcite{gupta2019document} for further details on training unsupervised DocNADE. 

Here, we extend the unsupervised version to DocNADE with a hybrid cost ${\mathcal L}^{hybrid}({\bf v})$, consisting of a (supervised) discriminative training cost $p(y=q | {\bf v})$ along with an unsupervised generative cost $p({\bf v})$ for a given query $q$ and associated document ${\bf v}$:
\begin{equation}
{\mathcal L}^{hybrid}({\bf v}) =  {\mathcal L}^{sup}({\bf v}) + \lambda \cdot {\mathcal L}^{unsup}({\bf v}) 
\end{equation}
where $\lambda \in [0, 1]$. The supervised cost is given by:
\[
{\mathcal L}^{sup}({\bf v})  = p(y=q|{\bf v}) = \mbox{softmax} ({\bf d} + {\bf S} \ {\bf h}({\bf v}))
\]    
Here, ${\bf S} \in \mathbb{R}^{L \times H}$ and ${\bf d} \in \mathbb{R}^{L}$ are output matrix and bias, $L$ is the total number of unique RDoC constructs (i.e., unique query labels). 

     {\bf Supervised Attention-based NTM (a-SupDocNADE)}: Observe in equation~\ref{eq:embeddingagrregation} that the DocNADE computes document representation ${\bf h}(v)$ via aggregation of word embedding vectors without considering attention over certain words. However, certain content words own high important, especially  in classification task. Therefore, we have introduced attention-based embedding aggregation in supDocNADE (Figure~\ref{fig:task1_pipeline},  left):     
\begin{equation}
{\bf h}({\bf v}) = f({\bf c} + \sum_{i=1}^D \alpha_i {\bf W}_{:,v_i})
\end{equation}
Here, $\alpha_i$ is an attention score of each word $i$ in the document ${\bf v}$, learned via supervised training. 

Additionally, we incorporate extra word features, such as  pre-trained word embeddings from several sources: FastText (${\bf E}^{fast}$) \cite{bojanowski2017enriching} and word2vec (${\bf E}^{word2vec}$) \cite{mikolov2013distributed}. 
We introduce these features by concatenating ${\bf h}_e ({\bf v})$ with ${\bf h}(\bf v)$ in the supervised portion of the a-supDocNADE model, as
\begin{equation}
{\bf h}_{e}({\bf v}) = f \Big({\bf c} + \sum_{i=1}^D \alpha_i ( {\bf E}_{:,v_i}^{fast} + {\bf E}_{:,v_i}^{word2vec})\Big)
\end{equation}

Therefore, the classification portion of a-supDocNADE with additional features is given by:
\[
p(q|{\bf v}) = \mbox{softmax}({\bf d} + {\bf S}{'} \cdot \mbox{concat}({\bf h}({\bf v}), {\bf h}_{e}({\bf v})))
\]    
where, ${\bf S}{'} \in \mathbb{R}^{H{'} \times L}$ and $H{'}=H+E^{fast}+E^{word2vec}$.

    \subsection{Traditional Methods for IR}
    {\bf BM25}:  A ranking function proposed by \newcite{robertson2009probabilistic} is used to estimate the relevance of a document for a given query. 
    
    {\bf BM25-Extra}:  The relevance score of BM-25 is combined with four extra features: 
    (1) percentage of query words with exact match in the document,
    (2) percentage of query words bigrams matched in the document, 
    (3) IDF weighted document vector for feature \#1, and 
    (4) IDF weighted document vector for feature \#2. 
Therefore, BM25-Extra returns a vector of 5  scores. 

    \subsection{System Description for RDoC Task-1}
    RDoC Task-1 aims at retrieving and ranking of PubMed abstracts (title and content) that are relevant for 8 RDoC constructs. Participants are provided with 8 clusters, each with a RDoC construct label and required to rank abstracts within each cluster based on their relevance to the corresponding cluster label. 
Each cluster contains abstracts relevant to its RDoC construct, while some (or most) of the abstracts are noisy in the sense that they belong to a different RDoC construct. Ideally, the participants are required to rank abstracts in each of the clusters by determining their relevance with the RDoC construct of the cluster in which they appear. 
  
  To address the RDoc Task-1, we learn a mapping function between latent representation ${\bf h}({\bf v})$ of a document (i.e.., abstract) ${\bf v}$ and its RDoC construct, i.e., query words $q$ in a supervised fashion. In doing so, we have employed supervised classifiers, especially supervised neural topic model {\bf a-supDocNADE} (section~\ref{sec:TM}) for document ranking.  We treat $q$ as label and maximize $p(q|{\bf v})$ leading to maximize ${\mathcal L}^{hybrid}({\bf v})$ in {\it a-supDocNADE} model.

As demonstrated in Figure~\ref{fig:task1_pipeline} (right), we perform document ranking in two steps: 

   (1) {\bf Document Relevance Ranking}: 
We build a supervised classifier using all the training documents and their corresponding labels (RDoC constructs), provided with the training set. At the test time, 
we compute prediction probability score $p(CID= q | {\bf v}^{test}(CID))$) of the label=CID for each test document ${\bf v}^{test}(CID)$ in the cluster, CID. This prediction probability (or confidence score) is treated as a relevance score of the document for the RDoC construct of the cluster.   
Figure~\ref{fig:task1_pipeline}(right) shows that we perform document ranking using the probability scores (col-2) of the RDoC construct (e.g. {\it loss}) within the cluster $C1$. Observe that a test document with least confidence for a cluster are ranked lower within the cluster and thus, improving mean average precision (mAP). Additionally, we also show the predicted RDoC construct in col-1 by the supervised classifier. 

   (2) {\bf Document Relevance Re-ranking}: Secondly, we re-ranked each document ${\bf v}$ (title+abstract) within each cluster (with label $q$) using  {\it unsupervised ranking}, where the relevance scores are computed as: 
(a) {\bf reRank(BM25-Extra)}: sum each of the 5 relevance scores to get the final relevance, and 
(b) {\bf reRank(QAR)}:  cosine-similarity(QAR(${\bf v}$), $\widetilde{\bf q}$).  

    \subsection{System Description for RDoC Task-2}
    The RDoC Task-2 aims at extracting the most relevant sentence from each of the PubMed abstract for the corresponding RDoC construct. Each abstract consists of title $t$ and sentences $s$ with an RDoC construct $q$.  

     To address RDoc Task-2, we first compute {\it multi-view} representation: BoW, TF-IDF and QAR (i.e., $\Phi_q(s^j)$) for each sentence $s^j$ in an abstract $d$. 
On other hand, we compute ESR representation for RDoC construct (query $q$) and title $t$ of the abstract $d$ to obtain $\widetilde{\bf q}$ and $\widetilde{\bf t}$, respectively. 
Figure~\ref{fig:task2_pipeline} and section~\ref{sec:textrepresentation} describe the computation of these representations. 
We then use the representations ($\Phi_q(s^j)$, $\widetilde{\bf t}$ and $\widetilde{\bf q}$) to compute a relevance scores of a sentence 
$s_j$ relative to $q$ and/or $t$ via {\it unsupervised} and {\it supervised} ranking schemes, discussed in the following section.  

    
    \subsubsection{Unsupervised Sentence Ranker} \label{section:USR}
     As shown in Figure~\ref{fig:task2_pipeline}, we first extract representations: $\Phi_q(s^j)$, $\widetilde{\bf t}$ and $\widetilde{\bf q}$ for the sentence $s^j$ query $q$ and title $t$. 
During ranking sentences within an abstract for the given RDoC construct $q$, we also consider title $t$ in computing the relevance score for each sentence relative to $q$ and $t$. 
It is inspired from the fact that the title often contains relevant terms (or words) appearing in sentence(s) of the document (or abstract).  
On top, we observe that $q$ is a very short text and non-descriptive, leading to minimal text overlap with $s$.  

We compute two relevance scores: $r_q$ and $r_t$ for a sentence $s^j$ with respect to a query $q$ and title $t$, respectively.
\[
r_q = sim(\widetilde{\bf q}, \Phi_q(s^j)) \mbox{ and } r_t = sim(\widetilde{\bf t}, \Phi_q(s^j))
\]

Now, we devise two ways to combine the relevance scores $r_q$ and $r_t$ in unsupervised paradigm:
\[
\mbox{version1:} \ r_1^{unsup} = r_q  \cdot r_q  +  r_t  \cdot r_t
\]
Observe that the relevance scores are weighted by itself. However, the task-2 expects a higher importance to the relevance score $r_q$ over $q_t$. 
Therefore, we coin the following weighting scheme to give higher importance to $r_q$ only if it is higher than $r_t$ otherwise we compute a weight factor $r'_t$ for $r_t$. 
\[
\mbox{version2:} \ r_2^{unsup} = r_q  \cdot r_q  +  r'_t  \cdot r_t
\]
where $r'_t$ is compute as:
\[
r'_t = (r_t > r_q) |r_t - r_q| 
\]
The relevance score  $r_2^{unsup}$ is effective in ranking sentences when a query and sentence does not overlap. In such a scenario, a sentence is scored by title, penalized by a factor of $|r_t - r_q|$.   

At the end, we obtain a final relevance score $r_f^{unsup}$ for a sentence $s^j$ by summing the relevance scores of BM25-Extra and $r_1^{unsup}$ or $ r_2^{unsup}$.

    \subsubsection{Supervised Sentence Ranker}\label{sec:supervisedRanker}
	
    Beyond unsupervised ranking, we further investigate sentence ranking in supervised paradigm by introducing a distance metric between the query (or title) and sentence vectors.  

    Figure~\ref{fig:task2_pipeline} describes the computation of relevance score for a sentence $s^j$ using a supervised sentence ranker scheme. Like the unsupervised ranker (section~\ref{section:USR}),  
the supervised ranker also employs vector representations: $\Phi_q(s^j)$, $\widetilde{\bf t}$ and $\widetilde{\bf q}$. Using the projection matrix $\bf G$, 
we then apply a projection to each of the representation to obtain $\Phi_q^p(s^j)$, $\widetilde{\bf t}^p$ and $\widetilde{\bf q}^p$. 
Here, the operator $\otimes$ performs concatenation of the projected vector with its input via residual connection. 
Next, we apply a Manhattan distance metric to compute similarity (or relevance) scores, following \newcite{gupta2018replicated}:   
\[
r^{sup}=\mbox{exp}\Big(-|| (\Phi^p_q(s^j), \widetilde{\bf q}^p) + \beta \ (\Phi^p_q(s^j), \widetilde{\bf t}^p)  ||_2\Big)
\]
where $\beta \in [0, 1]$ controls the relevance of title, determined by cross-validation. A final relevance score $r_f^{sup} \in [0, 1]$ is computed by feeding a vector [$r^{sup}$,  $r_{siamese}^{sup}$, BM25-extra] into a supervised linear regression, 
which is trained end-to-end by minimizing mean squared error between the $r_f^{sup}$ and \{0, 1\}, i.e., 1 when the sentence $s^j$ is relevant to query $q$. 
Here, $r_{siamese}^{sup}$ refers to a relevance score computed between $q$ and $s^j$ via Siamese-LSTM \cite{gupta2018replicated}.   

To perform sentence ranking within an abstract for a given RDoC construct $q$, the relevance score $r_f^{sup}$ (or $r_f^{unsup}$) is computed for all the sentences and a sentence with the highest score is extracted.

	\section{Experiments and Results}
	
	\begin{table}[t!]
		\begin{center}
			\resizebox{7.7cm}{!}{%
				\begin{tabular}{r|c|c|c|c|c|c|c|c||c}
					\multicolumn{1}{c|}{}                                                                     & \texttt{L1} & \texttt{L2} & \texttt{L3} & \texttt{L4} & \texttt{L5} & \texttt{L6} & \texttt{L7} & \texttt{L8} & \texttt{Total} \\ \hline
					\multicolumn{1}{r|}{\textit{All data}}                                                   & 39          & 38          & 47          & 21          & 28          & 27          & 48          & 18          & 266            \\ 
					\multicolumn{1}{r|}{\textit{Train set}}     & 31          & 30          & 37          & 16          & 22          & 21          & 38          & 14          & 209            \\ 
					\multicolumn{1}{r|}{\textit{Dev set}}  & 8           & 8           & 10          & 5           & 6           & 6           & 10          & 4           & 57             \\ 
					\multicolumn{1}{r|}{\textit{Test set (Task1)}} & 79          & 108         & 123         & 144         & 138         & 139         & 122         & 146         & 999            \\ 
					\multicolumn{1}{r|}{\textit{Test set (Task2)}} & 19          & 26          & 30          & 35          & 34          & 34          & 30          & 36          & 244            \\ 
			\end{tabular}}
		\end{center}
		\captionof{table}{Data statistics - \# of PubMed abstracts belonging to each RDoC construct in different data partitions. (\texttt{L1}: ``Acute Threat Fear''; \texttt{L2}: ``Arousal''; \texttt{L3}: ``Circadian Rhythms''; \texttt{L4}: ``Frustrative Nonreward''; \texttt{L5}: ``Loss''; \texttt{L6}: ``Potential Threat Anxiety''; \texttt{L7}: ``Sleep Wakefulness''; \texttt{L8}: ``Sustained Threat'')}
		\label{table:rdoc_tasks_dataset_statistics}
	\end{table}
	
	\subsection{Data Statistics and Experimental Setup}
	

	\begin{table}[t!]
		\begin{center}
			\resizebox{7.7cm}{!}{%
				\begin{tabular}{c|c|c||c}
					\textbf{Model}                                                                                      & \textbf{Feature}                                                    & \textbf{\begin{tabular}[c]{@{}c@{}}Classification\\ Accuracy\end{tabular}} & \textbf{\begin{tabular}[c]{@{}c@{}}Ranking\\ mAP\end{tabular}} \\ 
					\hline \hline
					\texttt{SVM}                                                                                        & BoW                                                                 &  0.947                                                                            &   \textbf{0.992}                                                               \\ \hdashline
					\multirow{3}{*}{\textit{\begin{tabular}[c]{@{}c@{}}Re-ranking\\ with\\ cluster label\end{tabular}}} & \textit{\texttt{reRank \#1}}                                                            &  -                                                                            &   \textbf{0.992}                                                               \\ 
					& \textit{\texttt{reRank \#2}}         &  -                                                                            &  \textbf{0.992}                                                                \\ 
					& \textit{\texttt{reRank \#3}} &  -                                                                            &  \textbf{0.992}                                                                \\ 
					\hline \hline
					\multirow{3}{*}{\texttt{a-supDocNADE}}              & Random init                                                         &  0.912                                                                            &   0.930                                                               \\ 
					& + FastText                                                          &  0.947                                                                           &  0.949                                                                \\ 
					& + BioNLP                                                            &  0.965                                                                            &  0.983                                                                \\ \hdashline
					\multirow{3}{*}{\textit{\begin{tabular}[c]{@{}c@{}}Re-ranking\\ with\\ cluster label\end{tabular}}} & \textit{\texttt{reRank \#1}}                                                             &  -                                                                            &   0.985                                                               \\ 
					& \textit{\texttt{reRank \#2}}         &  -                                                                            &  \textbf{0.994}                                                                \\ 
					& \textit{\texttt{reRank \#3}} &  -                                                                            &  \textbf{0.994}                                                                \\ 
			\end{tabular}}
		\end{center}
	\captionof{table}{RDoC Task-1 results (on development set): Classification accuracy and mean Average Precision (mAP) of \texttt{a-supDocNADE} and \texttt{SVM} models. Each model's \textit{classification accuracy} and \textit{ranking mAP} (using prediction probabilities) are shown together. Furthermore, each model's ranked clusters are re-ranked using different re-ranking algorithms. Best mAP score for each model is marked in \textbf{bold}. (\textit{\texttt{reRank \#1}}: ``reRank(BM25-Extra)''; \textit{\texttt{reRank \#2}}: ``reRank(QAR)''; \textit{\texttt{reRank \#3}}: ``reRank(BM25-Extra) + reRank(QAR)'')} \label{table:rdoc_task1_results}
	\end{table}
	
	{\bf Dataset Description}:
	Dataset for RDoC Tasks contains a total of 266 PubMed abstracts labelled with 8 RDoC constructs in a single label fashion.
	Number of abstracts for each RDoC construct is described in Table~\ref{table:rdoc_tasks_dataset_statistics}, where \textit{first} row describes the statistics for all abstracts and \textit{second} \& \textit{third} row shows the split of those abstracts into \textit{training} and \textit{development} sets maintaining a 80-20 ratio for each RDoC construct.
	For Task-1, each PubMed abstract contains its associated title, PubMed ID (PMID) and label (RDoC construct).
	In addition for Task-2, each PubMed abstract also contains a list of most relevant sentences from that abstract.
	Final evaluation test data for Task-1 \& Task-2 contains 999 \& 244 abstracts respectively.
	
	We use ``RegexpTokenizer'' from scikit-learn to tokenize abstracts and lower-cased all tokens.
	After this, we remove those tokens which occur in less than 3 abstracts and also remove stopwords (using nltk).
	For computing BM25-Extra relevance score, we use unprocessed raw text of sentences and titles.
	
	{\bf Experimental Setup}:
	As the training dataset labelled with RDoC constructs is very small, we use an external source of semantical knowledge by incorporating pretrained distributional word embeddings~\cite{zhang2019biowordvec} from FastText model~\cite{bojanowski2017enriching} trained on  the entire corpus of PubMed and MIMIC III Clinical notes~\cite{johnson2016mimic}.
	Similarly, we also use pretrained word embeddings \cite{moen2013distributional} from word2vec model~\cite{mikolov2013distributed} trained on PubMed and PMC abstracts.
	We create 3 folds\footnote{we only report results on $fold_1$ because of best scores on partial test dataset} of train/dev splits for cross-validation.
	
	
	{\bf RDoC Task-1}:
	For DocNADE topic model, we use latent representation of size 50. 
	We use pretrained FastText embeddings of size 300 and pretrained word2vec embeddings of size 200. 
	For SVM, we use Bag-of-words (BoW) representation of abstracts with radial basis kernel function.
	PubMed abstracts are provided in eight different clusters, one for each RDoC construct, for final test set evaluation.
	
	{\bf RDoC Task-2}:
	We use pretrained FastText embeddings to compute \textit{query-aware sentence representation} of a sentence ($\Phi_q(s^j)$), title ($\widetilde{\bf t}$) and query ($\widetilde{\bf q}$) representations.
	We also train Replicated-Siamese-LSTM~\cite{gupta2018replicated} model with input as sentence and query pair i.e., ($s^j, q$) and label as 1 if $s^j$ is relevant otherwise 0.
	We use $\beta \in \{0,1\}$.
	
	\begin{table}[t]
		\begin{center}
			\resizebox{7.7cm}{!}{%
				\begin{tabular}{c|c|c|c|c}
					\multicolumn{3}{c|}{\textbf{\begin{tabular}[c]{@{}c@{}}Ranking\\ (with Prediction Probability)\end{tabular}}} &
					\multicolumn{2}{c}{\textbf{\begin{tabular}[c]{@{}c@{}}Re-ranking\\ (with BM25-Extra)\end{tabular}}} \\ \cline{1-5} 
					\textbf{PMID}   & \textbf{Pred Prob} & \textbf{Gold Label} & \textbf{PMID}    & \textbf{Gold Label}                                 \\ \hline
					22906122                            &  0.90                                    &    PTA                                                                                        &   26005838                                             &    PTA                                                 \\ \hline
					24286750                            &  0.77                                    &    PTA                                                                                        &   22906122                                             &    PTA                                                 \\ \hline
					17598732                            &  0.61                                    &    PTA                                                                                        &   28828218                                             &    PTA                                                 \\ \hline
					26005838                            &  0.56                                    &    PTA                                                                                        &   26773206                                             &    PTA                                                 \\ \hline
					28316567                            &  0.46                                    &   \textit{Loss}                                                                                        &   24286750                                             &    PTA                                                 \\ \hline
					28828218                            &  0.45                                    &   PTA                                                                                         &   17598732                                             &    PTA                                                 \\ \hline
					26773206                            &  0.41                                    &   PTA                                                                                         &   28316567                                             &    \textit{Loss}                                                 \\ 
			\end{tabular}}
		\end{center}
		\captionof{table}{RDoC Task-1 analysis: Ranking of PubMed abstracts within ``Potential Threat Anxiety (PTA)'' cluster using supervised prediction probabilities ($p(q|\textbf{v})$). It shows that an intruder/noisy abstract (Gold Label: \textit{Loss}) is assigned higher probability than the abstracts with same Gold Label as the cluster. But, using re-ranking with BM25-Extra (\texttt{reRank(BM25-Extra)}) relevance score assigns lowest relevance to the intruder abstract.} \label{table:rdoc_task1_result_analysis}
	\end{table}
	
	\subsection{Results: RDoC Task-1}

	Table~\ref{table:rdoc_task1_results} shows the performance of supervised Document Ranker models i.e, \textbf{a-supDocNADE} and \textbf{SVM}, for Task-1.
	\textit{SVM} achieves a classification accuracy of 0.947 and mean average precision (mAP) of 0.992 by ranking the abstracts in their respective clusters using the supervised prediction probabilities ($p(q|\textbf{v})$).
	After that, we use three different relevance scores: (1) \textit{reRank(BM25-Extra)}, (2) \textit{reRank(QAR)} and (3) \textit{reRank(BM25-Extra)} + \textit{reRank(QAR)}, for re-ranking of the abstracts in their respective clusters.
	It is to be noted that the \textit{ranking mAP} of the clusters using prediction probabilities is already the best possible i.e., the intruder abstracts (abstracts with different label (RDoC construct) than the cluster label) are at the bottom of the ranked clusters.
	Therefore, re-ranking of these clusters would not achieve a better score.
	Similarly, we train \textit{a-supDocNADE} model with three different settings: (1) random weight initialization, (2) incorporating FastText embeddings ($\bf h_e(\bf v)$) and (3) incorporating FastText and word2vec embeddings ($\bf h_e(\bf v)$).
	By using the pretrained embeddings, the classification accuracy increases from 0.912 to 0.965, this shows that distributional pretrained embeddings carry significant semantic knowledge.
	Furthermore, re-ranking using \textit{reRank(BM25-Extra)} and \textit{reRank(QAR)} further results in the improvement of mAP score (0.994 vs 0.983) by shifting the intruder documents at the bottom of each impure cluster.
	
	\subsection{Analysis: RDoC Task-1}
	
	Table~\ref{table:rdoc_task1_result_analysis} shows an impure ``Potential Threat Anxiety'' cluster of abstracts containing an intruder abstract with label (RDoC construct) ``Loss''.
	When this cluster is ranked on the basis of prediction probabilities ($p(q|\bf v)$), then ``Loss'' abstract is ranked third from the bottom and it degrades the mAP score of the retrieval system.
	But after re-ranking this cluster using \textit{reRank(BM25-Extra)} relevance score, the ``Loss'' abstract is ranked at the bottom, thus maximizing the mAP score.
	Therefore, re-ranking with BM25-Extra on top of ranking with $p(q|\bf v)$ is, evidently, a robust abstract/document ranking technique.
	
	\subsection{Results: RDoC Task-2}
	
	\begin{table}[t!]
		\begin{center}
			\resizebox{7.7cm}{!}{%
				\begin{tabular}{c|c|c|c|c}
					\textbf{Model}                         & \textbf{Feature}                                                                                        & \textbf{Recall} & \textbf{F1} & \begin{tabular}[c]{@{}c@{}}\textbf{Macro-Average}\\\textbf{Accuracy}\end{tabular} \\ 
					\hline \hline
					\multirow{3}{*}{\textit{Unsupervised}} & reRank(BM25-Extra) [\#1]                                                                                              &  0.316                  &  0.387               &   0.631          \\ 
					& version1 [\#2]                                                                                                     &  \textbf{0.351}                  &  \textbf{0.412}               &  \textbf{0.701}           \\ 
					& version2 [\#3]                                                                                                     &  0.263                  &    0.345             & 0.526            \\ \hline
					\multirow{2}{*}{\textit{Supervised}}   & $r_f^{sup}$($\beta$ = 0) [\#4] &   \textbf{0.386}                 &    \textbf{0.436}             &     \textbf{0.772}        \\ 
					& $r_f^{sup}$($\beta$ = 1) [\#5] & 0.368                   &    0.424             &     0.737        \\ \hline
					\multirow{4}{*}{\textit{Ensemble}}     & \{\#1, \#2, \#4\}                                                                                                     &  \textbf{0.395}                  &  \textbf{0.441}               &   \textbf{0.789}          \\ 
					&  \{\#1, \#3, \#4\}                                                                                                    &  0.316                  &   0.387              &  0.631           \\ 
					&  \{\#2, \#4, \#5\}                                                                                                   &  \textbf{0.395}                  &   \textbf{0.441}              &    \textbf{0.789}         \\ 
					&  \{\#1, \#2, \#3, \#4, \#5\}                                                                                                   &  0.368                  &   0.424              &    0.737         \\ 
			\end{tabular}}
		\end{center}
		\captionof{table}{RDoC Task-2 results (on development set): Performance of unsupervised and supervised sentence rankers (Figure~\ref{fig:task2_pipeline}) under different configurations. Best scores for each model is marked in \textbf{bold}.} \label{table:rdoc_task2_results}
	\end{table}
	
	\begin{table*}[t!]
		\begin{center}
			\renewcommand*{\arraystretch}{1.15}
			\resizebox{16cm}{!}{%
				\begin{tabular}{l|cccccccc|c||cccccccc|c}
					\hline
					\multicolumn{1}{c}{\multirow{2}{*}{\textbf{Team}}}                                                   & \multicolumn{9}{|c||}{\textbf{RDoC Task-1 (Official Results)}}                                                                                  & \multicolumn{9}{c}{\textbf{RDoC Task-2 (Official Results)}}                                                                                  \\ \cline{2-19} 
					\multicolumn{1}{c|}{}                                                                                 & \texttt{L1}   & \texttt{L2}   & \texttt{L3}   & \texttt{L4}   & \texttt{L5}   & \texttt{L6}   & \texttt{L7}   & \texttt{L8}   & \texttt{mAP}  & \texttt{L1}   & \texttt{L2}   & \texttt{L3}   & \texttt{L4}   & \texttt{L5}   & \texttt{L6}   & \texttt{L7}   & \texttt{L8}   & \texttt{MAA}  \\ \hline
					\textit{MIC-CIS} & 0.85          & 0.92          & \textbf{1.00} & \textbf{0.73} & \textbf{0.78} & \textbf{0.94} & \textbf{1.00} & 0.63          & \textbf{0.86} & \textbf{0.74} & \textbf{0.73} & 0.47          & 0.37          & \textbf{0.74} & \textbf{0.59} & \textbf{0.60} & 0.42          & \textbf{0.58} \\ 
					\textit{Javad Rafiei Asl}                                                                              & \textbf{0.89} & \textbf{0.93} & \textbf{1.00} & 0.69          & 0.74          & 0.87          & \textbf{1.00} & \textbf{0.64} & 0.85          & 0.68          & 0.62          & 0.47          & 0.34          & 0.56          & 0.32          & 0.50          & 0.36          & 0.48          \\ 
					\textit{Ramya Tekumalla}                                                                               & 0.83          & 0.91          & \textbf{1.00} & 0.67          & 0.71          & 0.89          & \textbf{1.00} & \textbf{0.64} & 0.83          & 0.37          & 0.12          & 0.10          & 0.11          & 0.26          & 0.15          & 0.17          & 0.14          & 0.18          \\ 
					\textit{Daniel Laden}                                                                                  & 0.67          & 0.84          & \textbf{1.00} & 0.61          & 0.61          & 0.81          & \textbf{0.98} & 0.41          & 0.74          & -             & -             & -             & -             & -             & -             & -             & -             & -             \\ 
					\textit{Shyaman Jayasundara}                                                                           & -             & -             & -             & -             & -             & -             & -             & -             & -             & 0.47          & 0.42          & 0.60          & 0.29          & 0.62          & 0.38          & 0.57          & \textbf{0.47} & 0.48          \\ 
					\textit{Fei Li}                                                                                        & -             & -             & -             & -             & -             & -             & -             & -             & -             & 0.58          & 0.46          & \textbf{0.70} & \textbf{0.43} & 0.26          & 0.41          & 0.33          & \textbf{0.47} & 0.46          \\ \hline
			\end{tabular}}
		\end{center}
		\captionof{table}{RDoC Tasks official results - performance on test set of different competing systems. Best score in each column is marked in \textbf{bold}. (Refer to Table~\ref{table:rdoc_tasks_dataset_statistics} for header notations) (\texttt{mAP}: ``Mean Average Precision''; \texttt{MAA}: Macro-Average Accuracy)} \label{table:rdoc_tasks_official_results}
	\end{table*}
	
	\begin{table}[t]
		\begin{center}
			\resizebox{7.7cm}{!}{%
				\begin{tabular}{c|c|c}
					\multicolumn{3}{c}{\textbf{\begin{tabular}[c]{@{}c@{}}PubMed Abstract\\ (PMID: ``23386529''; RDoC construct: ``Loss'')\end{tabular}}}                                                                                                                                                                  \\ 
					\hline \hline
					\multicolumn{1}{c|}{\begin{tabular}[c]{@{}c@{}}\textit{Most Relevant Sentence}\\ (using \texttt{reRank(BM25-Extra)})\end{tabular}}                                           & \textit{\begin{tabular}[c]{@{}c@{}}Sentence\\ ID\end{tabular}} & \textit{\begin{tabular}[c]{@{}c@{}}Gold\\ Label\end{tabular}} \\ \hline
					\begin{tabular}[c]{@{}l@{}}Nurses are expected to care for \\ grieving women and families \\ suffering from perinatal loss.\end{tabular}                     & \#1                                                            & Not relevant                                                  \\ \hline \hline
					\multicolumn{1}{c|}{\begin{tabular}[c]{@{}c@{}}\textit{Most Relevant Sentence}\\ (using \texttt{Ensemble \{\#1, \#2, \#4\}})\end{tabular}}                           &       -                                                         &        -                                                       \\ \hline
					\begin{tabular}[c]{@{}l@{}}We found that nurses \\ experience a grieving process \\ similar to those directly suffering \\ from perinatal loss.\end{tabular} & \#6                                                            & Relevant                                                      \\ 
			\end{tabular}}
		\end{center}
		\captionof{table}{RDoC Task-2 analysis: This table shows that the most relevant sentence predicted using \texttt{reRank(BM25-Extra)} is actually not a relevant sentence, but \texttt{Ensemble \{\#1, \#2, \#4\}} (Table~\ref{table:rdoc_task2_results})  predicts the correct sentence as the most relevant.} 
		\label{table:rdoc_task2_analysis}
	\end{table}
	
	Table~\ref{table:rdoc_task2_results} shows results for Task-2 using \textit{three} unsupervised and \textit{two} supervised sentence ranker models.
	For unsupervised model, using \textit{reRank(BM25-Extra)} relevance score between a query ($q$), label (RDoC construct) of an abstract, and all the sentences ($s^j$) in an abstract, we get an macro-average accuracy (MAA) of 0.631.
	However, using \textit{version1} and \textit{version2} models (see Fig~\ref{fig:task2_pipeline}), we achieve a MAA score of 0.701 and 0.526 respectively.
	Higher accuracy of \textit{version1} model suggests that title ($t$) of an abstract also contains the essential information regarding the most relevant sentence.
	For supervised model, we get an MAA score of 0.772 and 0.737 by setting $\beta$ = 0 \& 1 in supervised relevance score ($r_f^{sup}$) equation in section~\ref{sec:supervisedRanker}.
	Hence, for supervised sentence ranker model, title ($t$) is playing a negative influence in correctly identifying the relevance ($r_f^{sup}$) of different sentences.
	Furthermore, we combine the knowledge of unsupervised and supervised sentence rankers by creating multiple ensembles (majority voting) of the predictions from different models.
	We achieve the highest MAA score of 0.789 by combining the predictions of (1) \textit{reRank(BM25-Extra)}, (2) \textit{version1}, and (3) $r_f^{sup}$ with $\beta = 0$.
	Notice that all the proposed supervised and unsupervised sentence ranking models (except [\#3]) outperform tranditional ranking models, e.g., \textit{reRank(BM25-Extra)} in terms of  query-document relevance score.

	\subsection{Analysis: RDoC Task-2}
	Table~\ref{table:rdoc_task2_analysis} shows that the most relevant sentence predicted by \textit{reRank(BM25-Extra)} is actually a non-relevant sentence.
	But an ensemble of predictions from both unsupervised and supervised ranker models correctly predicts the relevant sentence. This suggests that complementary knowledge of different models is able to capture the relevance of sentences on different scales and majority voting among them is, evidently, a robust sentence ranking technique.

	\subsection{Results: RDoC Task 1 \& 2 on Test set}
	
	Table~\ref{table:rdoc_tasks_official_results} shows the final evaluation scores of different competing systems for both the RDoC Task-1 \& Task-2 on final test set. Observe that our submission (MIC-CIS) scored a mAP score of $0.86$ and MAA of $0.58$ in Task-1 and Task-2, respectively. Notice that we outperform the second best system by $20.83\%$ ($0.58$ vs $0.48$) margin in Task2.

	\section{Conclusion}
	In conclusion, both supervised neural topic model and SVM can effectively perform ranking of PubMed abstracts in a given cluster based on the prediction probabilities. However, a further re-ranking using \textit{BM25-Extra} or \textit{query-aware sentence representation (QAR)} has proven to maximize the mAP score by correctly assigning the lowest relevance score to the intruder abstracts. Also, unsupervised and supervised sentence ranker models using query-title-sentence interactions outperform the traditional BM25-Extra based ranking model by a significant margin.
	
	In future, we would like to introduce complementary feature representation via hidden vectors of LSTM jointly with topic models and would like to further investigate the interpretability \cite{guptaMasterthesis:2015, DBLP:conf/emnlp/GuptaS18} of the proposed neural ranking models in the sense that one can extract salient patterns determining relationship between query and text.  Another promising direction would be introduce abstract information, such as part-of-speech and named entity tags \cite{DBLP:conf/naacl/LampleBSKD16, DBLP:conf/coling/GuptaSA16} to augment information retrieval (IR). 
	
	\section*{Acknowledgment}
	This research was supported by Bundeswirtschaftsministerium (bmwi.de), grant 01MD19003E (PLASS (plass.io)) at Siemens AG - CT Machine Intelligence, Munich Germany.

\bibliography{emnlp-ijcnlp-2019}
\bibliographystyle{acl_natbib}

\end{document}